# Space Narrative: Generating Images and 3D Scenes of Chinese Garden from Text using Deep Learning

Jiaxi Shi[1][0009-0006-9344-5894] and Hao Hua[1*][0000-0001-5988-7767]

[1] School of Architecture, Southeast University, Nanjing, China
whitegreen@seu.edu.cn

**Abstract.** The consistent mapping from poems to paintings is essential for the research and restoration of traditional Chinese gardens. But the lack of firsthand material is a great challenge to the reconstruction work. In this paper, we propose a method to generate garden paintings based on text descriptions using deep learning method. Our image-text pair dataset consists of more than one thousand Ming Dynasty Garden paintings and their inscriptions and postscripts. A latent text-to-image diffusion model learns the mapping from descriptive texts to garden paintings of the Ming Dynasty, and then the text description of Jichang Garden guides the model to generate new garden paintings. The cosine similarity between the guide text and the generated image is the evaluation criterion for the generated images. Our dataset is used to fine-tune the pre-trained diffusion model using Low-Rank Adaptation of Large Language Models (LoRA). We also transformed the generated images into a panorama and created a free-roam scene in Unity 3D. Our post-trained model is capable of generating garden images in the style of Ming Dynasty landscape paintings based on textual descriptions. The generated images are compatible with three-dimensional presentation in Unity 3D.

## 1 Introduction

In the context of ancient Chinese gardens, the narrative subject embodies a complex composition, incorporating multiple roles such as the creator, experiencer, and interpreter [1]. Based on the different identities and forms of works, creative endeavors can be broadly categorized into three types: lyrical poems expressing the sentiments of garden owners, articles detailing visitors' experiences, and visual depictions that recreate human sensory perceptions. Despite the varied narrative subjects, their creations converge upon the same object, namely, the abstract representation of tangible garden spaces. Hence, we can establish a three-dimensional perception of a specific garden space through the mutual corroboration and supplementation of the content presented from the perspectives of different narrative subjects in their works. Therefore, comparing information in garden paintings and poems has become an important method of analyzing and restoring traditional gardens [2]. For example, the materials survived from Ming and Qing dynasties of Jichang Garden (Wuxi, Jiangsu province) provided a reliable basis for the later reconstruction [3].

The traditional research methods highly rely on experience and professional knowledge based on long time research and practice. However, it is difficult to reconstruct the garden landscape that never appeared in any graphic material. In order to solve the problems of data scarcity of single garden restoration, we use a multimodal model to create the text-image correspondence of gardens in a particular period. By learning the correlation between graphic and textual data, the model can deduce garden images based on textual information.

## 2    Related Work

Comparing landscape painting of ancient Chinese literati and poems is an important method in garden verification and restoration. Due to the variety of firsthand materials and the complexity of operation, the traditional method is more suitable for single case rather than batch studies. Since 21th century, the digital technologies such as remote sensing, 3D reconstruction, and virtual reality have been widely employed for landscape heritage conservation [4,5]. But it is still difficult for these digital technologies to recreate unique historical landscapes in a specific period.

The generative machine learning model can generalise the character of training examples and create similar new data. In recent studies, pix2pix model is used to learn the topological relationship between the elements of the classic Chinese gardens and generate garden layout for a given site condition [6]. These methods focused on images and overlooked valuable text materials. Combining cross-referenced images and texts can provide a more reliable basis for studying classical gardens.

Machine learning models have been used to develop a method for analyzing Chinese garden design based on textual descriptions since 2003. The first research involved the conversion of garden element images into geometric shapes and the establishment of a relational diagram between garden spatial elements with descriptive text information [7]. Transformer and subsequent research efforts have established connections between images and text since 2017[8]. With DALL·E model one can generate semantically segmented garden plan according to the text describing classical gardens [9]. Diffusion models leverage iterative processes for more realistic and diverse image synthesis [10]. The pre-training models based on the latent text-to-image diffusion model, such as Stable Diffusion and Midjourney, can produce images which are highly consistent with the text semantics [11]. These works shed light on applying text-to-image models to generate garden landscapes from textual data, which can provide an efficient and easy-to-use method for the large batch of garden landscape research.

## 3    Method

This paper proposes a method of generating garden paintings based on textual descriptions using the latent text-to-image diffusion model. The overall process includes the following steps:

- Dataset creation: Collect the Ming Dynasty landscape paintings and related description texts and establish the image-text pair dataset.
- Pre-trained model fine-tuning: Input the image-text dataset into the pre-trained model and freeze the original model parameters. Use the LoRA method to fine-tune the attention processor in U-net.
- Testing and evaluation: Input the text describing Jichang Garden in the late Ming Dynasty to the fine-tuned model to guide it in generating landscape paintings, and carry out a similarity test on the generated results.
- VR representation: Select a group of generated results and create a walkable continuous scene in Unity 3D.

### 3.1 Text-Image Paired Dataset Creation

**Data Collection.** We first collected about 3000 landscape paintings and corresponding text description information from the Palace Museum, the National Palace Museum, Wuxi Museum, and Nanjing Museum through the Internet. Considering the impact of sample quality on the effect of machine learning, we filtered and modified the dataset according to the following conditions:

- The image should have clear architectural element.
- The image quality must reach six million pixels.
- For images that meet the above conditions but lack relevant descriptive text, the data will be discarded.

Based on the above conditions, we ultimately selected 1182 pairs of data and stored them as metadata in a format suitable for Diffusers Pipeline.

**Image Data Scaling.** Considering the pre-training model structure, we scale all images to 512×512 pixels. Table 1 below provides sample data.

**Table 1.** Sample Data

| Image | Metadata |
|---|---|
| 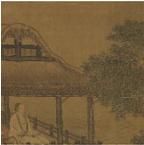 | {"file_name": " 000914N000000000.png", "additional_feature": " A figure reclines on a couch in the pavilion, whilst the figure in the boat plays on a flute and dangles his legs in the water. "} |
| 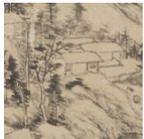 | {"file_name": " 000623N000000000.png", "additional_feature": " Cloud Forest Painting. To be clear and vulgar is to have nothing. The Xuanzai in this picture is accompanied by Dong Ju and Yunlin."} |

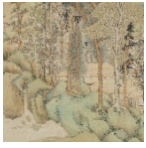 {"file_name": "000626N000000000.png", "additional_feature": " White clouds skirt the mountain; houses, where gentlemen sit talking leisurely, lie half-hidden by the thick forest. "}

### 3.2 Fine-tuning Latent Text-to-image Diffusion Model with LoRA

**Latent Text-to-image Diffusion Model Architecture.** The Latent text-to-image diffusion model is a machine learning system that gradually denoises random Gaussian noise in the latent space of lower dimensions until clear data (such as images) is obtained. The model is composed of three parts: the Variational Autoencoder (VAE) used for analyzing and generating images, the U-Net used for calculating noise residuals, and the CLIP text encoder used for converting text prompts into vectors in the latent space. The pre-training model actually used in this paper is stable-diffusion-v1-5 (512×512 resolution).

In the process of using the text-guided model to generate images, the model first converts each input text information into a 77x768 text embedding through the CLIP text encoder. Then, it takes the text embedding as a condition and uses U-Net to denoise the noise vector of the representative image, outputting the predicted noise residual. Afterward, the scheduler algorithm (in this paper, we use the preset PNDM scheduler) is employed to calculate the denoised image vector in the latent space based on the previous noise vector and the predicted noise residual. Finally, the VAE decoder is used to restore the vector representing the image to a full image of 512 pixels.

**Fine-tune with LoRA.** LoRA is a method to accelerate large-scale models' training with less memory consumption, which frees the trained model weights and inputs trainable rank decomposition matrix into each layer of the transformer architecture to reduce the number of trainable parameters. The total parameters of these rank decomposition matrices $|\Theta|$ are far less than the total parameters of the pre-training model $|\Phi_0|$. Parameters used during training $\Theta$ re-express the parameter increment $\Delta\Phi$ of the model as $\Delta\Phi = \Delta\Phi(\Theta)$ and constantly update the parameters in the rank-decomposition weight matrix to make the generated results similar to the data in the training set as much as possible. The optimization objectives are as follows:

$$\max_{\Theta} \sum_{(x,y)\in Z} \sum_{t=1}^{|y|} \log\bigl(p_{\Phi_0+\Delta\Phi(\Theta)}(y_t|x, y_{<t})\bigr) \qquad (1)$$

During the experiment, the parameters of the stable-diffusion-v1-5 (512x512 resolution) pre-training model were frozen. Only 32 layers of update matrix with rank 4 were added to the attention block of the UNet2DConditionalModel. Finally, the parameters of the low-rank update matrix were renewed on the self-made image-text pair dataset.

**Training and validation.** We choose the Diffusers library integrated with various pipelines in the Huggingface open-source machine learning community to fine-tune the model.

During the training process, the distance between the noise residual predicted by U-Net and the actual sample noise difference is recorded as the loss value (see Fig.1).

At the same time, we save the checkpoint model at the current time every 5000 training steps. We then generate four sample images using a textual prompt from the dataset in order to observe the training effect (see Fig.2). The cosine similarity between the generated images and the real image in the data pair is used to validate the accuracy of the results.

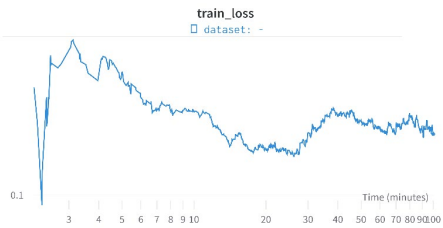 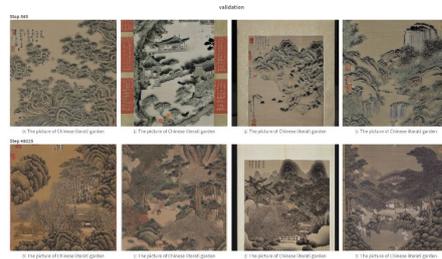

**Fig. 1.** Loss value     **Fig. 2.** Temporary results

### 3.3 VR Roaming Scene Representation

First, we use the functions integrated in the Diffusers library to enlarge and redraw a series of 512x512 late Ming Garden painting style Jichang Garden pictures generated by the fine-tuning model. Then, we use image processing software to combine multiple photos into a continuous image and merge them into a panorama. Finally, the panorama is made into a walkable continuous scene in Unity 3D.

## 4 Results

### 4.1 Jichang Garden Image Generation

**Landscape node image generation.** Wang Zhideng describes the path of travelling through the landscape around the square pond from the perspective of a tourist. This part of description consistent with four sequential painting in the "Fifty Scenes of Jichang Garden". A correspondence exists between them: It can be confirmed that the geometric forms of various landscape elements, their spatial relationships, and the sequence of sightseeing among these four scenes by reading and observing these materials.

We used Wang Zhideng's description to guide the model and generated four pictures (see Fig.3). The results indicated that the correspondence between the text and the garden images extracted by our model is fundamentally reasonable. It is capable of representing various spatial elements described in the text within the images. Compared to those researches using other models to generate semantic maps of gardens, our model's generated images not only depict the spatial relationships between various elements but also vividly showcase the scene intentions and stylistic features of late Ming Dynasty Chinese gardens.

However, it struggles to accurately generate a specific spatial element with a fixed geometric form in multiple tasks. These issues will be addressed in future research.

"泉自石隙泻沼中,声淙淙中琴瑟,临以屋,曰小憩。拾级而上,亭翼然峭倩青葱间者,为悬淙。引悬淙之流,甃为曲涧,茂林在上,清泉在下,奇峰秀石,含雾出云,于焉修禊,于焉浮杯,使兰亭不能独胜。曲涧水奔赴锦汇,曰飞泉,若出峡春流,盘涡飞沫,而后汪然渟然矣。"

Textual description in Wang Zhideng's "Notes on the Ji Chang Garden" [12].

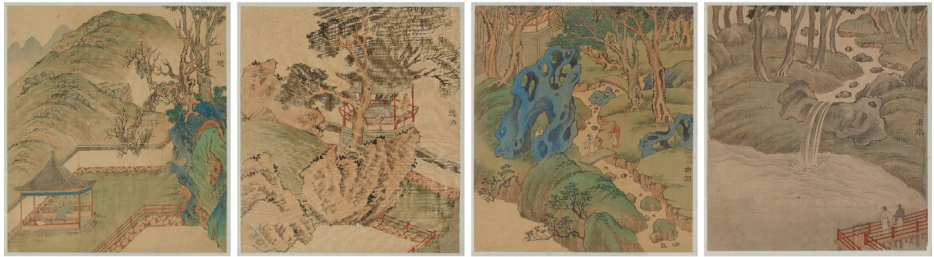

Paintings from "Fifty scenes of Jichang Garden": Xiao Qi, Xuan Cong, Qu Jian and Fei Quan [13].

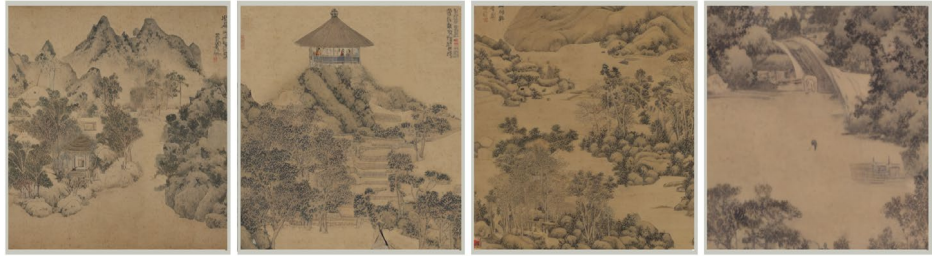

paintings generated by our model.

**Fig. 4.** Comparison of "Fifty Views of Jichang Garden" (middle) and the image generated by the model (below).

**Tour route image generation.** Although Wang Zhideng's narrative provides an account of his traversal as a visitor from one scene to another, the visual representation of the spatial connections between these scenes within the paintings remains non-intuitive. Therefore, we used the model's inpainting function to fill in the gaps between spatial sequences by synthesizing long scrolls, establishing a completely spatial perception (see Fig.5). Then the image is made into a panorama in the image processing software. With Unity 3D, one can immersively experience the scenes in the painting using virtual reality tools (see Fig.6).

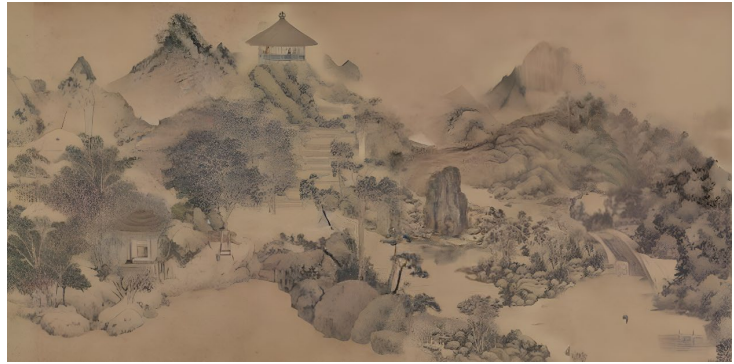

**Fig. 5.** A complete scene generated from four separate scene images using the inpainting function.

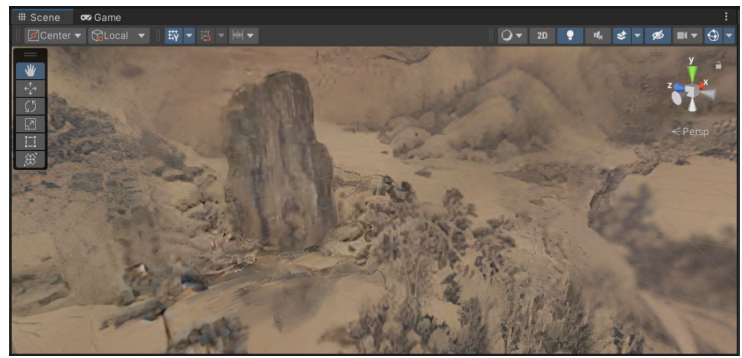

**Fig. 6.** A view of a roamable panorama created with Unity3D

## 5 Conclusion

This paper proposes a research method using a multimodal deep learning model to summarize the corresponding relationship between landscape painting and poetry and generate landscape images based on the prompt text. In this paper, we take Jichang Garden as an example to verify the model's feasibility in reproducing the landscape of Ming Dynasty. For constructing the dataset in this study, we intentionally incorporated works created by three types of narrative subjects within the context of Chinese gardens, aiming to comprehend garden spaces from the perspectives of different narrative subjects. The trained model is not only proficient at representing various garden elements in images but also adept at vividly showcasing garden styles characterized by distinct temporal and regional features.

However, due to scarce descriptions of spatial relationships in articles and poems, the generated images are not precise enough to depict positional relations as expected. Furthermore, due to differing horizons in model-generated images, segmenting the panorama into distinct ground and sky components poses challenges during the Unity 3D conversion process. Consequently, users may experience a subpar tour with the three-dimensional scene appearing as though they are floating in mid-air. One promising research direction involves identifying key spatial elements in the current image, establishing their geometric and semantic relationships using graph machine learning, and generating garden images with precise spatial orientation through natural language. Another area for exploration is enhancing the three-dimensional modeling process within Unity3D to create more realistic user experiences within the generated garden scenes, which has enormous potential for future applications in teaching and exhibition activities.